\documentclass{article}

\PassOptionsToPackage{numbers, compress}{natbib}
\usepackage[preprint]{nips_2018}

\usepackage[utf8]{inputenc} % allow utf-8 input
\usepackage[T1]{fontenc}    % use 8-bit T1 fonts
\usepackage{hyperref}       % hyperlinks
\usepackage{url}            % simple URL typesetting
\usepackage{booktabs}       % professional-quality tables
\usepackage{amsfonts}       % blackboard math symbols
\usepackage{nicefrac}       % compact symbols for 1/2, etc.
\usepackage{microtype}      % microtypography
\usepackage{wrapfig}

\usepackage{comment}
\usepackage{lipsum}

\usepackage{graphicx}
\usepackage{subcaption}
\usepackage{color} % pre-req for below macros
\usepackage{amsmath}

\usepackage{adjustbox}
\usepackage{tikz}
\usetikzlibrary{bayesnet, positioning,shapes.geometric, arrows, fit,calc}
\tikzstyle{block} = [draw, rectangle, fill=orange!50, text width=8em, text centered, minimum height=15mm, node distance=10em]
\tikzstyle{container} = [draw, rectangle, dashed, inner sep=0.7em]
\tikzstyle{arrow} = [thick,->,>=stealth]

\usepackage{"GrandMacros"}

\title{Adaptive and Calibrated Ensemble Learning with Dependent Tail-free Process}

\author{
  Jeremiah Zhe Liu$^{1}\quad$ John Paisley$^{2}\quad$ Marianthi-Anna Kioumourtzoglou$^{2}\quad$ Brent A. Coull$^{1}$ \\
  $^{1}$Harvard University, Cambridge, MA, 02139\\
  $^{2}$Columbia University, New York, NY, 10027\\
  \texttt{zhl112@mail.harvard.edu, jpaisley@columbia.edu} \\
  \texttt{mk3961@cumc.columbia.edu, bcoull@hsph.harvard.edu}
}

\begin{document}

\maketitle
\begin{comment}
\begin{abstract}
Ensemble learning is a mainstay in modern data science practice. Conventional ensemble algorithms assigns to base models a set of deterministic, constant model weights that (1) do not fully account for variations in base model accuracy across subgroups, nor (2) provide uncertainty estimates for the ensemble prediction, which could result in mis-calibrated predictions (i.e. precise but biased) that could in turn negatively impact the algorithm performance in real-word applications. In this work, we present an adaptive, probabilistic approach to ensemble learning using dependent tail-free process as ensemble weight prior. Given input feature $\bx$, our method optimally combines base models based on their predictive accuracy in the feature space $\bx \in \Xsc$, and provides interpretable uncertainty estimates both in model selection and in ensemble prediction. To encourage scalable and calibrated inference, we derive a structured variational inference algorithm that jointly minimize KL objective and the model's calibration score (i.e. Continuous Ranked Probability Score (CRPS)). We illustrate the utility of our method on both a synthetic nonlinear function regression task, and on the real-world application of spatiotemporal integration of particle pollution prediction models in New England.
\end{abstract}
\end{comment}

\vspace{-2 em}
\section{Introduction}
\vspace{-0.5em}

Ensemble learning is a mainstay in many modern machine-learning systems interacting with real-world data \citep{okun_ensembles_2011}. Conventional ensemble algorithms assign to base models a set of deterministic, constant model weights, not fully accounting for variability in the base model's  ability to capture different aspects of the data-generation mechanism, nor providing uncertainty estimates for the ensemble prediction. The motivating application for this work arises from the field of air pollution exposure assessment. To improve exposure assessment and minimize exposure measurement error, many different research groups are currently building highly resolved spatio-temporal prediction models. These different models have different inputs (from satellite remote sensing to chemical transport models and land use variables) and employ different algorithms (from linear mixed models, generalized additive models to neural networks).  Although the aim of each of these models is usually to maximize the global predictive accuracy, in practice this accuracy
varies both in space and in time (see, for example, Figure \ref{fig:pm25_base}). In this case, an ensemble algorithm assigning deterministic, spatially constant ensemble weights is neither accurate in prediction or informative in its prediction's reliability. Therefore for an ensemble method to be accurate and reliable in such  applications, it is crucial for the method to exhibit  \textit{\textbf{adaptivity}}, i.e. combine the base model predictions differently according their predictive performance over space and time, and also to provide a calibrated estimate of \textit{\textbf{uncertainty}}, in the sense that the model's predictive uncertainty faithfully reflects its actual likelihood of being correct. 
\begin{comment}
To design effective air pollution policies, the uncertainty in air pollution exposure assessment needs to be (1) comprehensively characterized, as it varies in space and time, in order to identify the areas with highest uncertainty, and (2) propagated in the estimation of health effects of air pollutants.
\end{comment}
 Recently, several ensemble approaches designed to improve spatio-temporal air pollution predictions have been developed, including hierarchical models with sophiscated covariance structure\citep{shaddick_data_2017, li_constrained_2017}, bootstrap aggregation of regression trees or neural networks \citep{li_ensemble_2017, hu_estimating_2017, wang_deep_2018}, and stacked generalization of multiple black-box algorithms\citep{xiao_ensemble_2018}. However, to our knowledge, no method to date has integrated spatio-temporal weighing of the base models, i.e. has assign larger weights at each space and time point to the base model with the highest accuracy. Importantly, no method has provided comprehensive intra- and inter-model characterization of the spatio-temporal uncertainty in the predictions.

To address the above gaps in methodology, in this work we present a feature-adaptive, probabilistic approach to ensemble learning  using a dependent tail-free process as the ensemble weight prior. Specifically, we model the ensemble weight as a random measure $\mu: \Fsc \times \Xsc \rightarrow [0, 1]$ that depends on the input feature $\bx$ (e.g. the spatial location), where $\hat{f} \in \Fsc$ is the space of base models and $\bx \in \Xsc$ is the input feature space. Therefore, as $\bx \in \Xsc$ changes, $\mu(\hat{f}, \bx)$ distributes the weights differently among the base models, while at
the same time quantifying uncertainty in weight assignment due to the fact that we take these weights as random measures. The resulting method incorporates the prior information about the base models, and provides interpretable uncertainty quantification that can be decomposed into uncertainty due to model selection and that due to ensemble prediction. 

To scale computation to the real-world sized problems without sacrificing the model's ability to accurately quantify uncertainty, in Section \ref{sec:inference}, we develop a novel variational inference algorithm that orients the variational family toward both approximating the model posterior, and also toward producing a \textit{calibrated} predictive distribution, in the sense that the predictive distribution should be consistent with the empirical distribution of the data\citep{gneiting_probabilistic_2007}. We achieve this by defining the objective function to be a composition of two distance measures: a Kullback-Leibler (KL) distance that measures the variational family's quality in approximating the model posterior, and an additional "calibration" distance that measures the variational predictive distribution's consistency with the empirical distribution of the data. The specific calibration distance we consider is the Cramer-von Mises (CvM) distance \citep{anderson_distribution_1962}, i.e. the $L_2$ distance between the model's predictive cumulative distribution function (CDF) $\hat{P}(y_i < t|\bx_i)$ and the data's empirical CDF: $CvM(\hat{P}, y_i) = 
\int_t [ \hat{P}(y_i < t|\bx_i) - I(y_i < t) ]^2 dt$. As distance measures between probability distributions, both KL and CvM induce quality measures for model's uncertainty quantification known as  \textit{proper scoring rules} \citep{gneiting_strictly_2007}. Compared to the KL distance, the CvM distance comprehensively access the entire predictive function's quality in approximating the emprical distribution. Consequently, the \textit{Continuously Ranked Probability Score} (i.e. the scoring rule induced by the CvM distance) is commonly preferred as a more robust alternative to logarithm score (i.e. the scoring rule induced by the KL)\citep{selten_axiomatic_1998, gneiting_probabilistic_2007}. In Section \ref{sec:exp}, we empirically investigate the behavior of our method and compare its performance with that of the traditional ensemble methods on a nonlinear function regression benchmark. We conclude with a real-world application of spatial integration of $PM_{2.5}$ pollution prediction models in the greater Boston area.

\vspace{-0.5em}
\section{Model}
\vspace{-0.5em}

\label{sec:model}
\begin{wrapfigure}{r}{4cm}
\centering
\begin{tikzpicture}[thick,scale=0.8, every node/.style={scale=0.8}]
% nodes
     \node[const] (k_u) {$k_{\mu}$};
     \node[const, right = of k_u] (lambda) {$\lambda$};     
     \node[latent, below = 0.5cm of k_u] (g) {$\bg_k$};
     \node[latent, below = 0.5cm of g] (mu) {$\bmu$};        
     \node[obs, right = of mu] (f_hat) {$\hat{f}_k$};  
     \node[latent, left = of mu] (eps) {$\bepsilon$};                                                      
     \node[const, above = 1.65cm of eps] (k_e) {$k_{\epsilon}$};                
     \node[latent, below = of mu] (f_ens) {$f$};                                     
     \node[latent, left = of f_ens] (sigma) {$\sigma$};                        
     \node[obs, below = 0.5cm of f_ens] (y) {$y$};                                             
% plate
	{
	\plate [inner sep=0.7cm, xshift=0.02cm, yshift=0.2cm]
	{plate_k}{(g) (mu) (f_hat)} {$k = 1, \dots, K$};
	}
% edge
	\edge [->] {k_u} {g} 
	\edge [->] {lambda} {g} 
	\edge [->] {k_e} {eps} 	
	\edge [->] {g} {mu} 
	\edge [->] {mu, f_hat, eps} {f_ens}
	\edge [->] {f_ens, sigma} {y}	
\end{tikzpicture}
\caption{Graphical representation of the proposed model.}
\label{fig:graphic_model}
\vspace{-4em}
\end{wrapfigure}
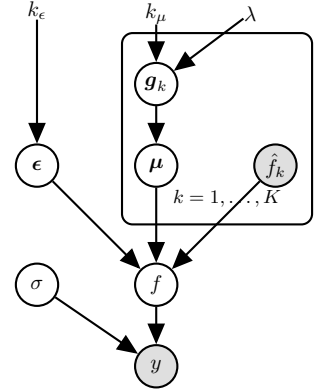

Given observations $\{\bx_i, y_i\}_{i=1}^N$ and a set of base models $\Fsc = \{\hat{f}_k\}_{k=1}^K$, we assume there exists a ensemble function $f$ such that:
\begin{align*}
y_i | f, \bx_i &\sim N(f(\bx_i), \sigma^2), \quad
f(\bx_i) = \sum_{k=1}^K \hat{f}_k(\bx_i)\bmu(\hat{f}_k, \bx_i) + \bepsilon(\bx_i),\\
\bmu \sim & \; \mbox{Tailfree} (\Pi, \Gsc, \Lambda), \quad
\bepsilon \sim \Gsc\Psc(0, k_\epsilon).
%\label{eq:f_sum}
\end{align*}
Here $\bmu: \Fsc \times \Xsc \rightarrow [0, 1]$ is a random measure that controls the contribution of each individual base model $\hat{f}_k$ to the overall ensemble, depending on the location in the feature space $\bx \in \Xsc$. It follows a dependent tail-free process which is constructed from a collection of normalized Gaussian processes $g \in \Gsc$ and sparse-inducing hyperparameters $\lambda \in \Lambda$ (See the end of this section for details). The residual process $\bepsilon$ is a flexible Gaussian process that captures the systematic bias shared by the base prediction functions. In terms of uncertainty quantification, $\bmu$ and $\bepsilon$ play distinct roles in the decomposition of  the overall uncertainty into that due to \textit{model selection} and that due to \textit{prediction}. Specifically, the posterior uncertainty in $\bmu(\hat{f}_k, \bx)$ reflects uncertainty in \textit{model selection}, which is expected be high when model predictions disagree and there are few observations to justify confident selection, and is expected to be low otherwise. On the other hand, the posterior uncertainty in $\bepsilon(\bx)$ reflects uncertainty in prediction, which describes the ensemble's additional uncertainty in its final prediction, conditional on model selection (See, e.g. Figure \ref{fig:unc_decomp}).\\
\textbf{Dependent Tail-free Process}
In the simplest scenario, given a set of $K$ prediction functions $\Fsc=\{\hat{f}_k\}_{k=1}^K$ with no grouping structure among them, we can consider $\Fsc$ as a depth 1 tree with $K$ leaf nodes (we denote this tree as $\Pi$), and model the ensemble weights $\bmu(\hat{f}_k, \bx)$ as $P(\hat{f}_k | \mbox{\texttt{root}}, \bx)$, i.e. the conditional probability of the $k^{th}$ leaf node given its parent (i.e. the root node, see Appendix Figure \ref{fig:partition}). Specifically, we model $P(\hat{f}_k | \mbox{\texttt{root}}, \bx)$ as the softmax transformation of $K$ independent Gaussian processes corresponding to each $\hat{f}_k$:
\begin{align}
P(\hat{f}_k | \texttt{root}, \bx) &= \frac{exp \big( g_{k}(\bx)/\lambda_{\texttt{root}} \big)}{ \sum_{k'=1}^{K} exp \big( g_{k}(\bx)/\lambda_{\texttt{root}} \big)}, 
\qquad
\{ g_k \}_{k=1}^K \stackrel{iid}{\sim} \Gsc\Psc(\bzero, k_{\mu}).
\label{eq:cond_prob}
\end{align}
Here $\lambda_{\texttt{root}}$ is the temperature parameter that controls the sparsity in model selection among the children of $\texttt{root}$. Denoting the collection of Gaussian processes $\{ g_k \}_{k=1}^K$ corresponding to each node in the tree $\Pi$ as $\Gsc$, and the collection of all temperature parameters in $\Pi$ as $\Lambda$, we have specified a dependent tail-free process (DTFP) prior for the ensemble weights $\bmu \sim \mbox{Tailfree}(\Pi, \Gsc, \Lambda)$ \cite{jara_class_2011}. Figure \ref{fig:graphic_model} shows a graphical representation of our model in this simple scenario. In the more common scenario where models exhibit grouping structure, e.g. some $\hat{f}_k$ in the ensemble come from the same model family or are trained on the same dataset, we can incorporate this group structure by organizing $\hat{f}_k$'s into a corresponding tree structure, and parition the model space $\Fsc$ accordingly (see Appendix Figure \ref{fig:partition}). Modelling for $\bmu$ under such recursive partitioning scheme can also be carried out naturally using the tail-free process \citep{ferguson_prior_1974, jara_class_2011}, where the probability of each leaf node is modelled as a sequence of independent conditional probabilities in its ancestry:
%\begin{align}
$
\bmu(\hat{f}_k, \bx) = 
P \big( \hat{f}_k \big|\, {\small \texttt{Parent}}(\hat{f}_k), \bx \big) * 
\Big[ 
\prod_{f \in {\small \texttt{Anc}}(\hat{f}_k)} 
P \big(f \big|\, {\small \texttt{Parent}}(f), \bx \big)
\Big],
$
%\label{eq:tail_free}
%\end{align}
where we have denoted $\texttt{Anc}(\hat{f}_k)$ as the set of $f_k$'s ancestors in the tree, and denoted $\texttt{Parent}(f)$ the immediate parent of a node $f$. The conditional probabilities $P \big(f \big|\, {\small \texttt{Parent}}(f), \bx \big)$ are modelled similarly as in (\ref{eq:cond_prob}), but with $g$'s corresponding to the siblings of $f$. In this case, the tail-free construction allows differential sparsity in model selection within each model group (by estimating different temperature parameter within each group), and provides information on the importance of the group of models at each level of the hierarchy.

\vspace{-0.5em}
\section{Variational Inference}
\vspace{-0.5em}
\label{sec:inference}

Denoting $\bz = \{\Gsc, \Lambda, \bepsilon, \sigma\}$ as the collection of all model variables, and $q_{\btheta}(\bz)$ as the variational family indexed by parameter $\btheta$, our variational objective of interest is the composition of the KL distance between the variational posterior and the model posterior, and the CvM distance between the variational predictive CDF and the empirical CDF of the data, i.e.
\begin{align}
\Lsc \big( \btheta \big| \, \bz, \{\bx_i, y_i\} \big) = 
KL \Big[q_{\btheta}(\bz) \big|\big| p(\bz|\bx_i, y_i) \Big] + 
CvM \Big[ \hat{P}_{\btheta}, y_i \Big],
\label{eq:vi_obj}
\end{align}
where $\hat{P}_{\btheta}(y<t|\bx_i) = \int_{-\infty}^t E_{q_{\btheta}(\bz)} \big(p(y|\bz, \bx_i)\big) dy$ is the variational predictive CDF. To perform variational learning with respect to $\Lsc$, we minimize the KL distance with respect to its negative evidence lower bound (ELBO) following the standard practice \citep{blei_variational_2016}, and perform Monte Carlo gradient update on the CvM distance. Specifically, we express the CvM distance as the expectation over samples from variational posterior \citep{gneiting_strictly_2007}:
$E( |\by - y_i| ) - \frac{1}{2} E( |\by - \by'| )$, where $
\by, \by' \stackrel{i.i.d.}{\sim} E_{q_{\btheta}(\bz)}[ p(y|\bz, \bx_i)]$, and derive its unbiased gradient estimator using the standard score gradient method \cite{paisley_variational_2012, ranganath_black_2013}. Variance reduction techniques such as Rao-Blackwellization  \cite{ranganath_black_2013} can be applied to encourage stable convergence. We use sparse Gaussian process \citep{titsias_variational_2009} as the variational family for $\Gsc$ and $\bepsilon$, and fully-factored lognormals for $\Lambda$ and $\sigma$. (See Appendix \ref{sec:gradient} for detail)

\vspace{-0.5em}
\section{Experiment and Application}
\label{sec:exp}
\vspace{-0.5em}
\subsection{Nonlinear function approximation}
\label{sec:exp_1d}
\vspace{-0.5em}

\begin{figure}[!ht]
\centering
\scalebox{.8}{
  \includegraphics[height=.23\linewidth, width=.35\linewidth]{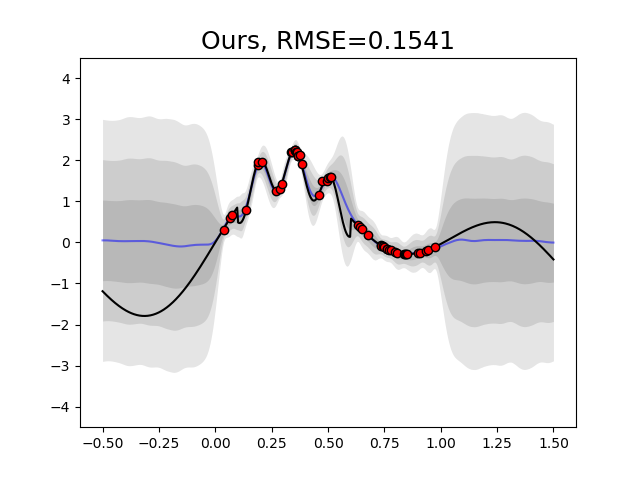}
  \includegraphics[height=.23\linewidth, width=.35\linewidth]{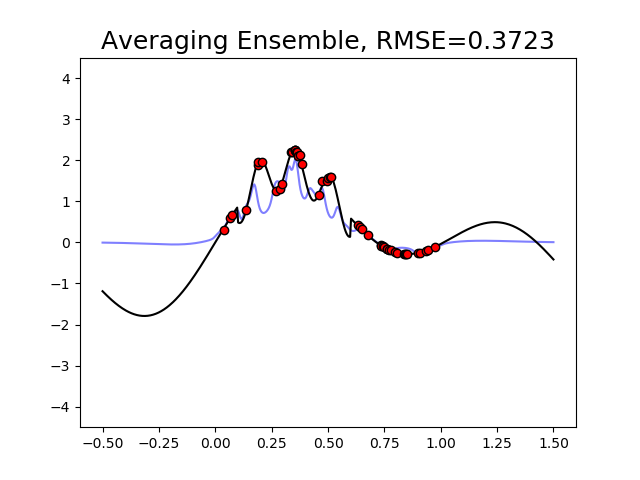}
  \includegraphics[height=.23\linewidth, width=.35\linewidth]{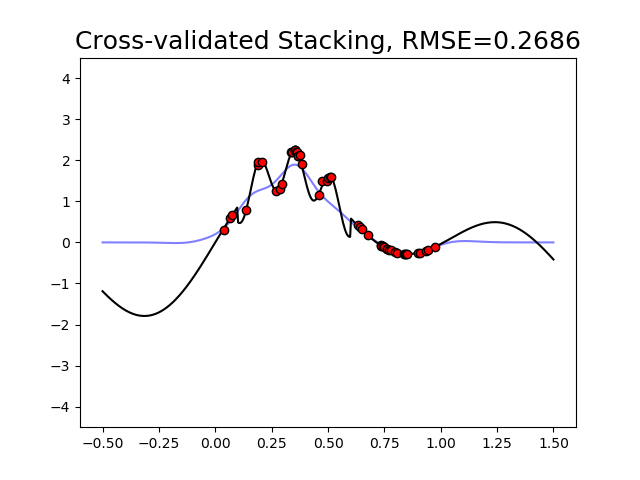}
}

\scalebox{.8}{
  \includegraphics[height=.23\linewidth, width=.35\linewidth]{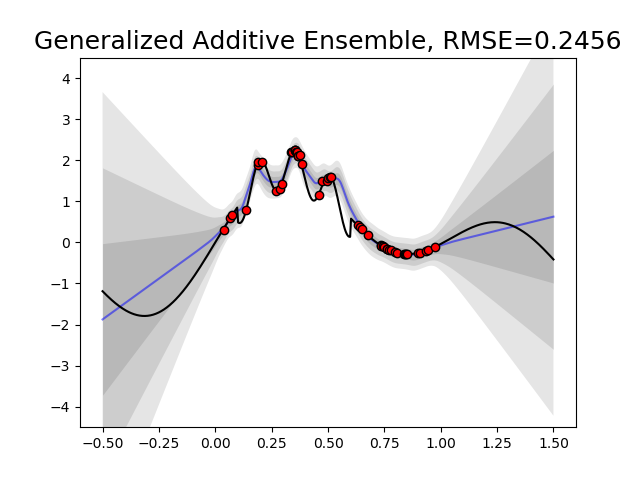}
  \includegraphics[height=.23\linewidth, width=.35\linewidth]{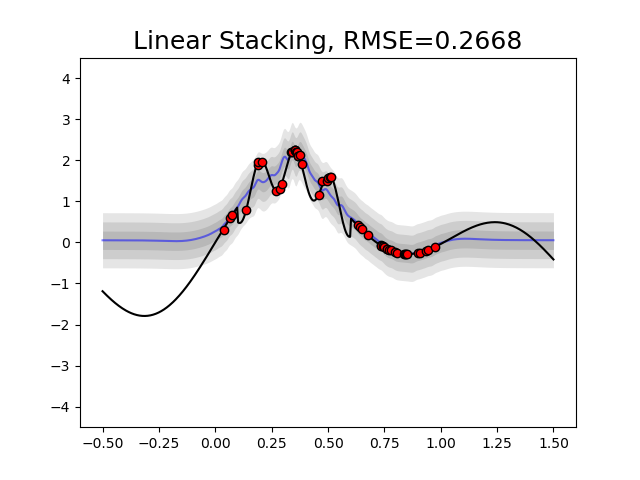}
  \includegraphics[height=.23\linewidth, width=.35\linewidth]{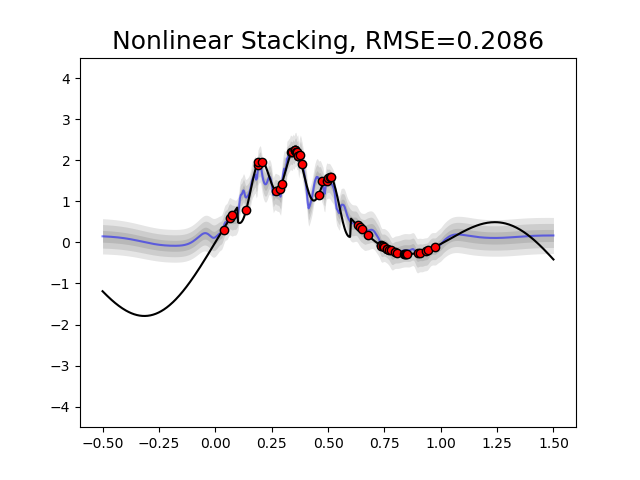}  
  }
 \caption{Comparison in prediction and uncertainty quantification of different ensemble methods. The grey bands indicating models' $68\%$, $95\%$ and $99\%$ predictive interval.}
  \label{fig:exp_1d_comp}
\end{figure}

We first investigate the model's behavior in prediction and uncertainty quantification on a 1-D nonlinear regression task. We compare the performance of our method with five other commonly used ensemble algorithms: (1) \textbf{avg} is the averaging ensemble that simply average over base model predictions, (2) \textbf{cv-stack} is the stacked generalization method \citep{breiman_stacked_1996} that aggregates model using simplex weight by minimizing their combined cross-validation errors. 
(3) \textbf{lnr-stack} and (4) \textbf{nlr-stack} are the linear and the nonlinear stacking methods that train a linear regression model or an additive B-spline regression \citep{wahba_spline_1990} using the base-model predictions as features, and finally (5) \textbf{gam} is the generalized additive ensemble \citep{xiao_ensemble_2018} that linearly combines the base-model prediction, and uses an extra smoothing spline term to mitigate any systematic bias. While \textbf{avg} and \textbf{cv-stack} produce only deterministic predictions, \textbf{lnr-stack}, \textbf{nlr-stack} and \textbf{gam} provide predictive distributions for the outcome.

\vspace{-0.5em}
\begin{table}[!ht]
\centering
\scalebox{.9}{
\begin{tabular}{|c|c|c|c|}
\hline\hline
Model & Ours & \textbf{avg} & \textbf{cv-stack} \\
\hline
Validation RMSE & $0.1531 \pm 0.017$ & $0.3723 \pm 0.028$ & $0.2686 \pm 0.014$ 
\\
\hline\hline
Model & \textbf{gam} & \textbf{lnr-stack} & \textbf{nlr-stack} \\
\hline
Validation RMSE & $0.2463 \pm 0.018$ & $0.2623 \pm 0.011$ & $0.2086 \pm 0.012$  
\\
\hline\hline
\end{tabular}
}
\caption{Mean and Standard Deviation for validation RMSE in 1-D regression task}
\label{tb:exp_1d_res}
\vspace{-1.5em}
\end{table}

Detailed experiment protocol is available in Appendix  \ref{sec:exp_1d_proto}. Briefly, we generate data from the composition of a \textit{global} function representing the slow-varying, global trend and a \textit{local} function representing the fast-varying, local fluctuations. We train four kernel regression models using RBF kernel with length-scale parameters $\{0.1, 0.2, 0.01, 0.02\}$. As a result, none of the base models can fit the data-generation mechanism universally well across $x \in (0, 1)$ (see Figure \ref{fig:base_exp_1d}). We train the six ensemble models on seperately generated 20 holdout observations, and evaluate each ensemble model's RMSE on a validation set of 500 observations. We repeat above training and evaluation procedure 100 times, and report the mean and standard deviation of the validation RMSE in Table \ref{tb:exp_1d_res}. We also visualize the models' behavior in prediction in one such training-evaluation instance in Figure \ref{fig:exp_1d_comp}. As shown, comparing the mean prediction (blue line), \textbf{avg}, \textbf{cv-stack} and \textbf{lnr-stack} produce either overly smooth or overly complex fits, due to assigning constant weights to the base models. On the other hand, \textbf{nlr-stack} and \textbf{gam} produce closer fit to the data-generation mechanism.
%, relying on either nonlinear transformation of the base models, or the extra spline "residual" term that mitigates the systematic bias in the linear ensemble. However, both models 
But they tend to overfit the observations in the holdout dataset, producing either unnecessary local fluctuations (\textbf{nlr-stack}), or extrapolating improperly in regions outside $x \in (0, 1)$ (\textbf{gam}), resulting in higher validation RMSE compared to that of our model. In comparison, our model produces smooth fit that closely matches the data-generation mechanism in regions where holdout observation is available or the model agreement is high, and produces smooth interpolation with high uncertainty in regions with few holdout observations and model agreement is low (e.g. $x \in (0.5, 0.75)$), indicating proper quantification of uncertainty. Examining the predictive intervals of other ensemble methods, we find that these intervals tend to vary less flexibly within the range of $x \in (0, 1)$, sometimes failing to reflect the increased uncertainty in regions where the data is sparse and the model agreement is low, and resulting in overly narrow confidence intervals (e.g. \textbf{gam} in $x \in (0.5, 0.75)$). We also quantitatively assess these models' quality in uncertainty quantification (in terms of the true coverage probability of the $p\%$ predictive intervals for $p \in (0, 1]$) in Appendix \ref{sec:exp_1d_unc_quant}.

\subsection{Spatial integration of air pollution predictions in New England region}

%Ambient air pollution has long being recognized as a global environmental threat to human health. 
In air pollution assessment, many research groups are developing distinct spatio-temporal models (exposure models) to predict ambient air pollution exposures of study participants even in areas where air pollution monitors are sparse. Depending on the prediction model and the inputs, disagreement among model predictions are consistently observed across space and time (Figure \ref{fig:pm25_base}), and information on prediction uncertainty are generally unavailable, leading to difficulties in exposure assessment for downstream health effect investigations. Here we use our ensemble method to aggregate the spatiotemporal predictions of three state-of-the-art $PM_{2.5}$ exposure models (\citep{kloog_new_2014, di_hybrid_2016, donkelaar_global_2016} in Figure \ref{fig:pm25_weight}) to produce a coherent set of spatiotemporal exposure estimate, along with information on predictive uncertainty. We perform our ensemble framework on the base models' out-of-sample prediction for 43 monitors across greater Boston area during year 2011. 
%We used RBF kernels for the ensemble weight $\bmu$, and the Ornstein–Uhlenbeck process kernel for the residual process, following standard practice in spatial modelling literature \citep{wispelaere_air_1984}. The length-scale parameters for both kernels are estimated through grid search based on minimizing the leave-one-out cross-validation error. 
We report the ensemble methods leave-one-out RMSE in Table \ref{tb:app_comp}, and visualize our model's posterior prediction and uncertainty in Figure \ref{fig:pm25_pred}, and the estimated ensemble weights in Appendix Figure \ref{fig:pm25_weight}. 
As shown, we observed elevated uncertainty close to Brockton region (where the base models disagrees) and regions further away from metro area (where the monitors are sparse), reflecting uncertainty in model selection and prediction that is consistent with empirical evidence.

\begin{table}[!ht]
\centering
\scalebox{.9}{
\begin{tabular}{|c|c|c|c|}
\hline\hline
Model & Ours & \textbf{avg} & \textbf{cv-stack} \\
\hline
loo RMSE & $0.7580 \pm 0.0883$ & $1.6768 \pm 0.124$ & $1.5437 \pm 0.1275$ 
\\
\hline\hline
Model & \textbf{gam} & \textbf{lnr-stack} & \textbf{nlr-stack} \\
\hline
loo RMSE & $1.0771 \pm 0.1566$ & $1.1626 \pm 0.1421$ & $1.2327 \pm 0.1265$  
\\
\hline\hline
\end{tabular}
}
\caption{Mean and Standard Deviation for leave-one-out RMSE in annual $PM_{2.5}$ ensemble prediction}
\label{tb:app_comp}
\vspace{-2em}
\end{table}

\clearpage

\begin{comment}
\section{Conclusion}
\begin{enumerate}
\item Future work
\begin{enumerate}
\item interrogate theoretical performance: prior support and posterior concentration \citep{yang_minimax_2014}.
\item experiment variational inference with :
\begin{enumerate}
\item different divergence (Wasserstein \citep{ambrogioni_wasserstein_2018}, $\chi^2$ distance \citep{dieng_variational_2016},  Renyi diveregence \citep{li_renyi_2016}), and different scoring rules. Evaluate difference in performance.
\item more flexible variational family, e.g. variational program\citep{ranganath_operator_2016} , NICE or NVP.
\end{enumerate}
\item flexible alternative to gaussian process. deep kernel learning \citep{wilson_deep_2016}? neural process \citep{garnelo_neural_2018}?
\end{enumerate}
\end{enumerate}

\subsubsection*{Acknowledgments}
Funding. Discussion (Stephan Hoyer, kernel scoring rule. Lorenzo Trippa, discussion on formulation). Software support TensorFlow Probability team.
\end{comment}

\newpage

\subsubsection*{Acknowledgments}
Authors would like to thank members of the Google Accelerated Science team for helpful comments and discussions, and especially Stephan Hoyer for introducing the concept of proper scoring rule.

\bibliographystyle{abbrv}
\bibliography{report}

\begin{thebibliography}{10}

\bibitem{anderson_distribution_1962}
T.~W. Anderson.
\newblock On the {Distribution} of the {Two}-{Sample} {Cramer}-von {Mises}
  {Criterion}.
\newblock {\em The Annals of Mathematical Statistics}, 33(3):1148--1159, Sept.
  1962.

\bibitem{blei_variational_2016}
D.~M. Blei, A.~Kucukelbir, and J.~D. McAuliffe.
\newblock Variational {Inference}: {A} {Review} for {Statisticians}.
\newblock {\em arXiv:1601.00670 [cs, stat]}, Jan. 2016.
\newblock arXiv: 1601.00670.

\bibitem{breiman_stacked_1996}
L.~Breiman.
\newblock Stacked regressions.
\newblock {\em Machine Learning}, 24(1):49--64, July 1996.

\bibitem{di_hybrid_2016}
Q.~Di, P.~Koutrakis, and J.~Schwartz.
\newblock A hybrid prediction model for {PM}2.5 mass and components using a
  chemical transport model and land use regression.
\newblock {\em Atmospheric Environment}, 131:390--399, Apr. 2016.

\bibitem{donkelaar_global_2016}
A.~v. Donkelaar, R.~V. Martin, M.~Brauer, N.~C. Hsu, R.~A. Kahn, R.~C. Levy,
  A.~Lyapustin, A.~M. Sayer, and D.~M. Winker.
\newblock Global {Estimates} of {Fine} {Particulate} {Matter} using a
  {Combined} {Geophysical}-{Statistical} {Method} with {Information} from
  {Satellites}, {Models}, and {Monitors}.
\newblock Mar. 2016.

\bibitem{ferguson_prior_1974}
T.~S. Ferguson.
\newblock Prior {Distributions} on {Spaces} of {Probability} {Measures}.
\newblock {\em The Annals of Statistics}, 2(4):615--629, July 1974.

\bibitem{gneiting_probabilistic_2007}
T.~Gneiting, F.~Balabdaoui, and A.~E. Raftery.
\newblock Probabilistic forecasts, calibration and sharpness.
\newblock {\em Journal of the Royal Statistical Society: Series B (Statistical
  Methodology)}, 69(2):243--268, Apr. 2007.

\bibitem{gneiting_strictly_2007}
T.~Gneiting and A.~E. Raftery.
\newblock Strictly {Proper} {Scoring} {Rules}, {Prediction}, and {Estimation}.
\newblock {\em Journal of the American Statistical Association},
  102(477):359--378, Mar. 2007.

\bibitem{hu_estimating_2017}
X.~Hu, J.~H. Belle, X.~Meng, A.~Wildani, L.~A. Waller, M.~J. Strickland, and
  Y.~Liu.
\newblock Estimating {PM}2.5 {Concentrations} in the {Conterminous} {United}
  {States} {Using} the {Random} {Forest} {Approach}.
\newblock {\em Environmental Science \& Technology}, 51(12):6936--6944, June
  2017.

\bibitem{jara_class_2011}
A.~Jara and T.~E. Hanson.
\newblock A class of mixtures of dependent tail-free processes.
\newblock {\em Biometrika}, 98(3):553--566, Sept. 2011.

\bibitem{kloog_new_2014}
I.~Kloog, A.~A. Chudnovsky, A.~C. Just, F.~Nordio, P.~Koutrakis, B.~A. Coull,
  A.~Lyapustin, Y.~Wang, and J.~Schwartz.
\newblock A new hybrid spatio-temporal model for estimating daily multi-year
  {PM}2.5 concentrations across northeastern {USA} using high resolution
  aerosol optical depth data.
\newblock {\em Atmospheric Environment}, 95:581--590, Oct. 2014.

\bibitem{li_constrained_2017}
L.~Li, F.~Lurmann, R.~Habre, R.~Urman, E.~Rappaport, B.~Ritz, J.-C. Chen, F.~D.
  Gilliland, and J.~Wu.
\newblock Constrained {Mixed}-{Effect} {Models} with {Ensemble} {Learning} for
  {Prediction} of {Nitrogen} {Oxides} {Concentrations} at {High}
  {Spatiotemporal} {Resolution}.
\newblock {\em Environmental Science \& Technology}, 51(17):9920--9929, Sept.
  2017.

\bibitem{li_ensemble_2017}
L.~Li, J.~Zhang, W.~Qiu, J.~Wang, and Y.~Fang.
\newblock An {Ensemble} {Spatiotemporal} {Model} for {Predicting} {PM}2.5
  {Concentrations}.
\newblock {\em International Journal of Environmental Research and Public
  Health}, 14(5), May 2017.

\bibitem{okun_ensembles_2011}
O.~Okun, G.~Valentini, and M.~Re.
\newblock {\em Ensembles in {Machine} {Learning} {Applications}}.
\newblock Springer, Sept. 2011.
\newblock Google-Books-ID: ZqcMBwAAQBAJ.

\bibitem{paisley_variational_2012}
J.~Paisley, D.~Blei, and M.~Jordan.
\newblock Variational {Bayesian} {Inference} with {Stochastic} {Search}.
\newblock {\em arXiv:1206.6430 [cs, stat]}, June 2012.
\newblock arXiv: 1206.6430.

\bibitem{ranganath_black_2013}
R.~Ranganath, S.~Gerrish, and D.~M. Blei.
\newblock Black {Box} {Variational} {Inference}.
\newblock {\em arXiv:1401.0118 [cs, stat]}, Dec. 2013.
\newblock arXiv: 1401.0118.

\bibitem{selten_axiomatic_1998}
R.~Selten.
\newblock Axiomatic {Characterization} of the {Quadratic} {Scoring} {Rule}.
\newblock {\em Experimental Economics}, 1(1):43--61, June 1998.

\bibitem{shaddick_data_2017}
G.~Shaddick, M.~L. Thomas, A.~Green, M.~Brauer, A.~Donkelaar, R.~Burnett, H.~H.
  Chang, A.~Cohen, R.~V. Dingenen, C.~Dora, S.~Gumy, Y.~Liu, R.~Martin, L.~A.
  Waller, J.~West, J.~V. Zidek, and A.~Prüss-Ustün.
\newblock Data integration model for air quality: a hierarchical approach to
  the global estimation of exposures to ambient air pollution.
\newblock {\em Journal of the Royal Statistical Society: Series C (Applied
  Statistics)}, 67(1):231--253, June 2017.

\bibitem{titsias_variational_2009}
M.~Titsias.
\newblock Variational {Learning} of {Inducing} {Variables} in {Sparse}
  {Gaussian} {Processes}.
\newblock In {\em Artificial {Intelligence} and {Statistics}}, pages 567--574,
  Apr. 2009.

\bibitem{wahba_spline_1990}
G.~Wahba.
\newblock {\em Spline {Models} for {Observational} {Data}}.
\newblock SIAM, Sept. 1990.
\newblock Google-Books-ID: ScRQJEETs0EC.

\bibitem{wang_deep_2018}
J.~Wang and G.~Song.
\newblock A {Deep} {Spatial}-{Temporal} {Ensemble} {Model} for {Air} {Quality}
  {Prediction}.
\newblock {\em Neurocomputing}, 314:198--206, Nov. 2018.

\bibitem{xiao_ensemble_2018}
Q.~Xiao, H.~H. Chang, G.~Geng, and Y.~Liu.
\newblock An ensemble machine-learning model to predict historical {PM}2.5
  concentrations in {China} from satellite data.
\newblock {\em Environmental Science \& Technology}, Oct. 2018.

\end{thebibliography}

\newpage
\appendix
\renewcommand\thefigure{\thesection.\arabic{figure}}    
\setcounter{figure}{0}    

\section{Additional Figures for Model Description}

\begin{figure}[!ht]
\centering
  \includegraphics[width=.3\linewidth]{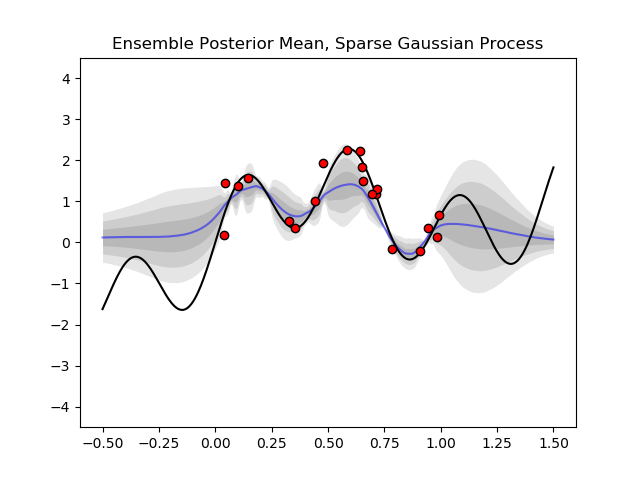}
  \includegraphics[width=.3\linewidth]{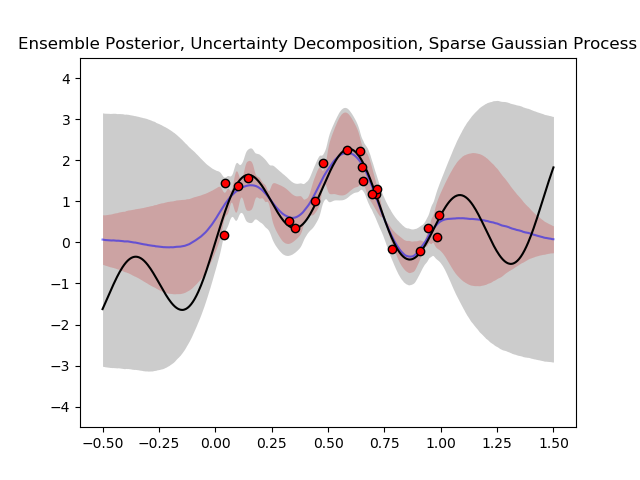}
  
\caption{Different components of predictive uncertainty decomposed by our model.\\ \textbf{(left)} Ensemble prediction and uncertainty estimate without residual process; \\
\textbf{(right)} Decomposition of uncertainty in model selection (red) and in prediction (grey).}
\label{fig:unc_decomp}
\end{figure}

\begin{figure}[ht]
\centering
\resizebox{0.8\textwidth}{!}{
    \begin{subfigure}[b]{0.45\textwidth}
        \centering
        \includegraphics[width=\linewidth]{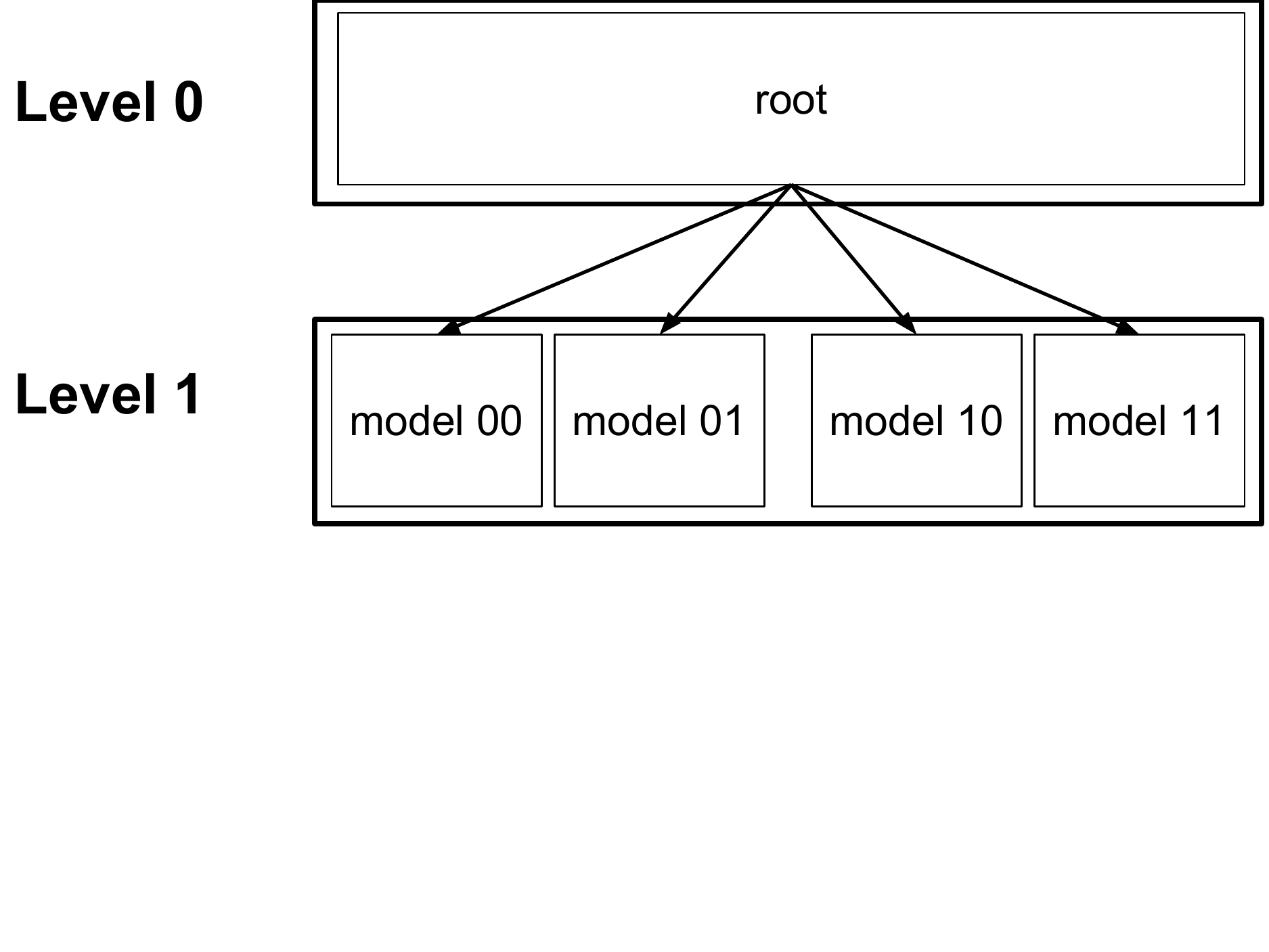}
        \caption{A naive partition}
    \end{subfigure}%
    \hspace{10mm}
    \begin{subfigure}[b]{0.45\textwidth}
        \centering
        \includegraphics[width=\linewidth]{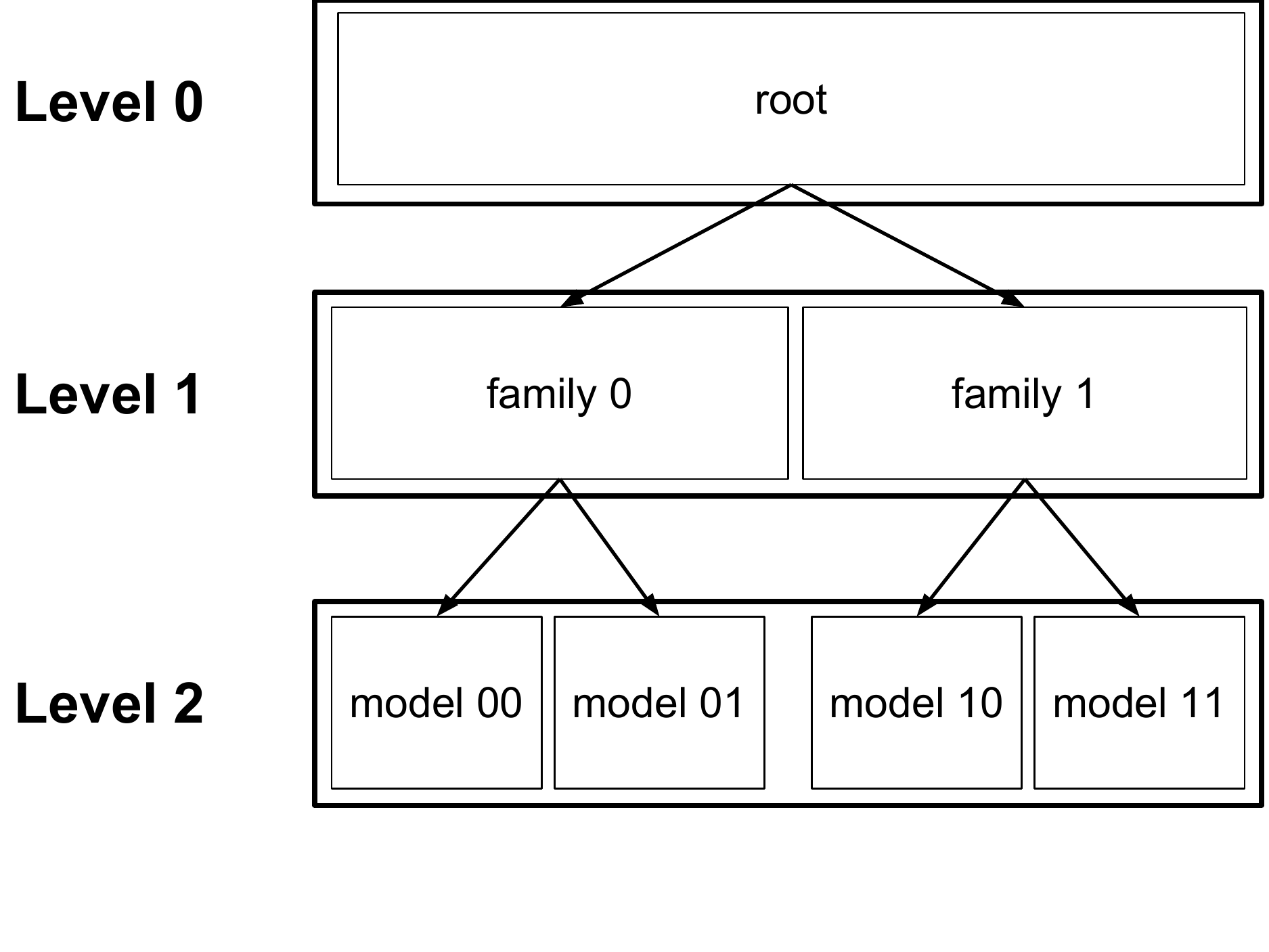}
        \caption{A partition utilizing model family information.}
    \end{subfigure}%
    }
\caption{An example partition of the model space $\Fsc$.
In the naive partition on the left, \\
$\bmu(f_{\texttt{model}_{00}}, \bx)=P(f_{\texttt{model}_{00}}| \; \texttt{root}, \bx)$, and in the right, \\
$\bmu(f_{\texttt{model}_{00}}, \bx) = P(f_{\texttt{model}_{00}}|f_{\texttt{family}_0}, \bx)P(f_{\texttt{family}_{0}}| \; \texttt{root}, \bx)$
}
\label{fig:partition}
\vspace{-10pt}
\end{figure}

In the right figure, the two conditional probabilities are modelled as:
\begin{align*}
P(f_{\texttt{model}_{00}}|f_{\texttt{family}_0}, \bx) &= exp(g_{\texttt{model}_{00}}/\lambda_{\texttt{family}_0} )/
\big[\sum_{i=0}^1 
exp( g_{\texttt{model}_{0i}}/\lambda_{\texttt{family}_0})\big] \\
P(f_{\texttt{family}_{0}}|\texttt{root}, \bx) &= exp( g_{\texttt{family}_{0}}/\lambda_{\texttt{root}})/ \big[ \sum_{i=0}^1  exp(g_{\texttt{family}_{i}})/\lambda_{\texttt{root}}\big]
\end{align*}

\clearpage
\section{Expression for Gradient of the Variational Inference Objective}
\label{sec:gradient}

\textbf{Variational family} Due to our focus on reliable uncertainty quantification, we find the naive mean-field approximation with fully-factored Gaussians tend to under-estimate predictive uncertainty, and produces non-smooth predictions that overfits the observation (see Appendix Figure \ref{fig:vi_result}). Consequently, we adopt a structured approximation based on sparse Gaussian process \citep{titsias_variational_2009}. Specifically, we factor the variational family into independent groups of Gaussian processes $\{\bepsilon, \Gsc \}$ and variance/temperature parameters $\{\sigma, \Lambda \}$ as 
$q(\bz) = 
q(\bepsilon) * \big( \prod_{g \in \Gsc} q(g) \big)
* \Big[
q(\sigma) * \big( \prod_{\lambda \in \Lambda} q(\lambda) \big)
\Big]$,
where we model $q(\bepsilon)$ and $q(g)$'s using sparse Gaussian processes, and model $q(\sigma)$ and $q(\lambda)$'s using fully factored log normal distributions.

\begin{figure}[ht]
\centering
  \includegraphics[width=0.33\linewidth]{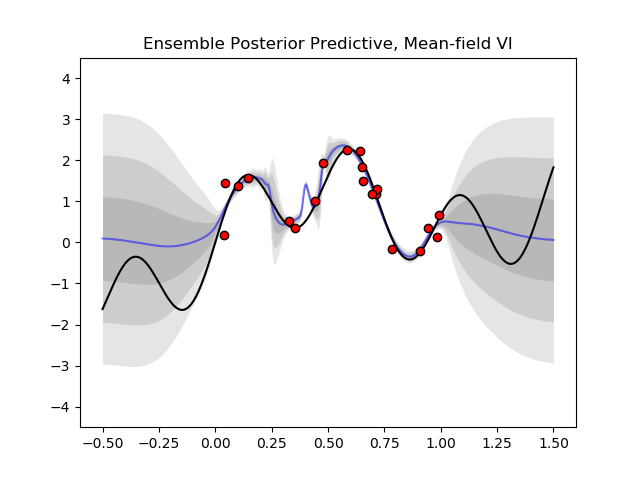}
  \includegraphics[width=0.33\linewidth]{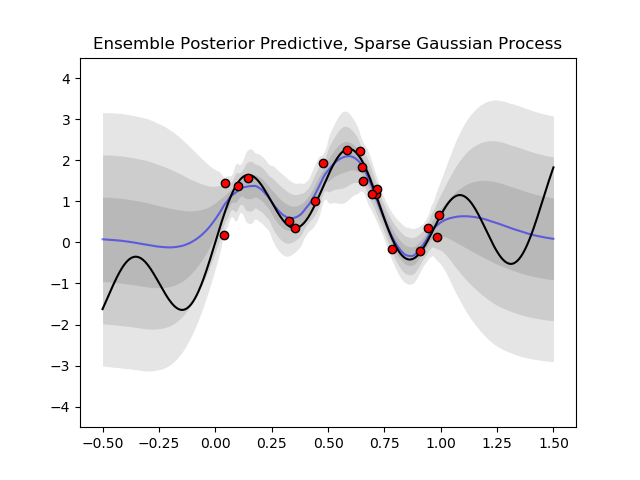}
\caption{Comparison of posterior predictive mean and uncertainty (95\% credible interval) between fully-factored mean-field VI \textbf{(left)} and structured VI based on sparse Gaussian process \textbf{(right)}.}
\label{fig:vi_result}
\end{figure}

\textbf{Gradient of the Variational Inference Objective}
Denote $q_{\btheta}(\bz)$ the variational distribution and $p_{\btheta}(\by|\bx)$ as the variational posterior predictive distribution (i.e.  $p_{\btheta}(\by|\bx)=\int p(\by|\bz, \bx) q_{\btheta}(\bz) d\bz$), then:
\begin{align*}
\nabla_{\btheta} KL(q_{\btheta} || p) &= 
E_{q_{\btheta}} \Big( (log \, p(\bz, \bx) - log \, q_{\btheta}(\bz)) * \nabla_{\btheta} log \, q_{\btheta}(\bz)  \Big)
\\
\nabla_{\btheta} CRPS(q_{\btheta}, p | \bx) &= 
E_{p_{\btheta}}\Big(
g(\by, y_{obs}) * \nabla_{\btheta} log \, p_{\btheta}(\by|\bx) 
\Big) - \\
& \qquad E_{p_{\btheta}} \Big(
g(\by, \by') * \frac{\nabla_{\btheta} log \, p_{\btheta}(\by|\bx) +  \nabla_{\btheta} log \, p_{\btheta}(\by'|\bx)}{2} 
\Big)
\end{align*}
where the gradient for predictive log likelihood $log \, p_{\btheta}(\by|\bx)$ can be written as:
\begin{align*}
\nabla_{\btheta} log \, p_{\btheta}(\by) &= 
\frac{E_{q_{\btheta}} \big( p(\bx|\bz) * \nabla_{\btheta} log \, q_{\btheta}(\bz) \big)}
{E_{q_{\btheta}} \big(  p(\bx|\bz) \big) }
\end{align*}

\clearpage
\section{Experiment Protocol and Additional Result for the 1-D Regression}

\subsection{Experiment Protocol}
\label{sec:exp_1d_proto}
\begin{figure}[!ht]
\centering
  \includegraphics[width=0.5\linewidth]{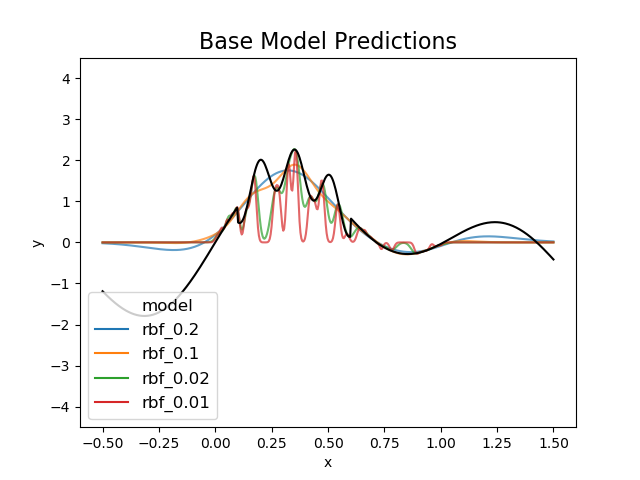}
  \caption{Data generation function and the deterministic predictions from base models in the 1D experiment. 
  \textbf{Black Line}: data-generation function. \textbf{Colored Line}: Base model predictions.}
	\label{fig:base_exp_1d}  
\end{figure}

To prepare base models in the ensemble, we randomly generate 20 data points $x_i \sim Uniform(0, 1)$, and generate $y_i = f(x_i + \epsilon_i)$, where $\epsilon_i \stackrel{i.i.d.}{\sim} N(0, 0.01)$ and the data-generating function $f(x)=f_{\mbox{\tiny \tt slow}}(x) + f_{\mbox{\tiny \tt fast}}(x)$ is the composition of a smooth, slow-varying global function $f_{\mbox{\tiny \tt slow}}(x)=x + sin(4*x)+sin(13*x)$ on $x \in (-0.5, 1.5)$, and a fast-varying, local function $f_{\mbox{\tiny \tt fast}}(x)=0.5 * sin(40*x)$ on $x \in (0.1, 0.6)$ (black lines in Figure \ref{fig:exp_1d_comp}). We train four different kernel regression models on separetely generated datasets, using Radial Basis Function (RBF) kernel with two groups of length-scale parameters ${\tt smooth}=\{0.2, 0.1\}$ and ${\tt complex}=\{0.02, 0.01\}$ to represent two groups of models with different smoothness assumptions, resulting in $\Fsc = \{\hat{f}_k\}_{k=1}^4$. As shown in Figure \ref{fig:base_exp_1d}, no base model can predict the ground truth universally well across $x \in (0, 1)$.  We then train all the ensemble methods on a holdout dataset of 20 data points generated using the same mechanism. For our model, we use RBF kernel for both $\bmu$ and $\bepsilon$, where we put prior $LogNormal(-1., 1.)$ on the  RBF's length-scale parameters so they are estimated automatically through the inference procedure. The spline hyperparameters for \textbf{nlr-stack} and \textbf{gam} are selected using random grid search over $10^3$ candidates based on model's cross-validation error.

After training, we evaluate each model's RMSE on a validation dataset of 500 data points spaced evenly between $x \in (0, 1)$. We repeat above training and evaluation procedure 100 times on randomly generated holdout datasets, and report the mean and standard deviation of the validation RMSE in Table \ref{tb:exp_1d_res}. We also visualize the models' behavior in prediction of one such training-evaluation instance in Figure \ref{fig:exp_1d_comp}.

\subsection{Quantitative Assessment of Uncertainty Quantification}
\label{sec:exp_1d_unc_quant}

\begin{figure}[!ht]
\centering
  \includegraphics[width=.33\linewidth]{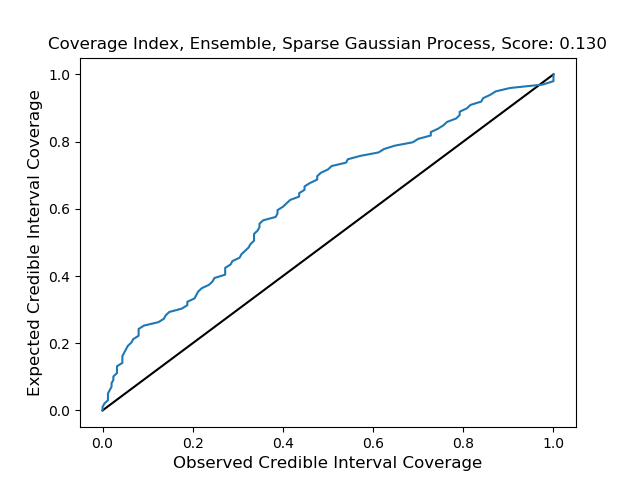}
  \includegraphics[width=.33\linewidth]{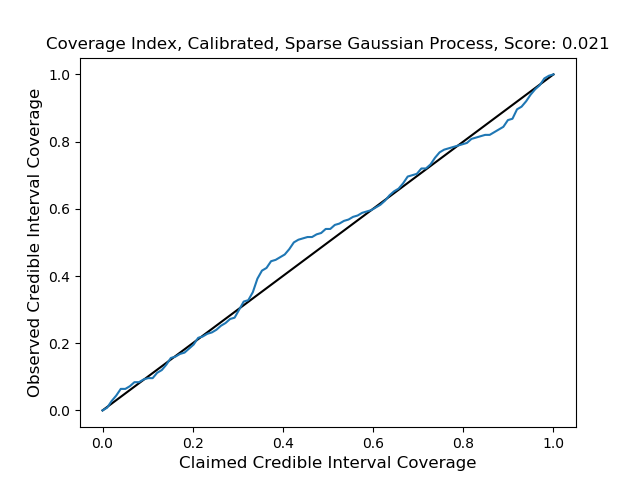}

  \includegraphics[width=.32\linewidth]{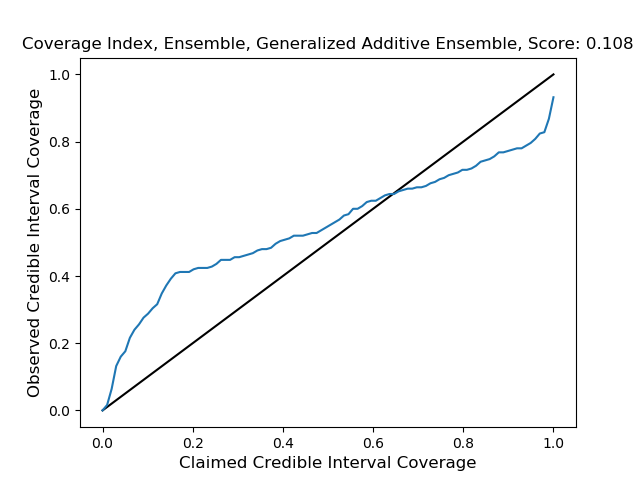}
  \includegraphics[width=.32\linewidth]{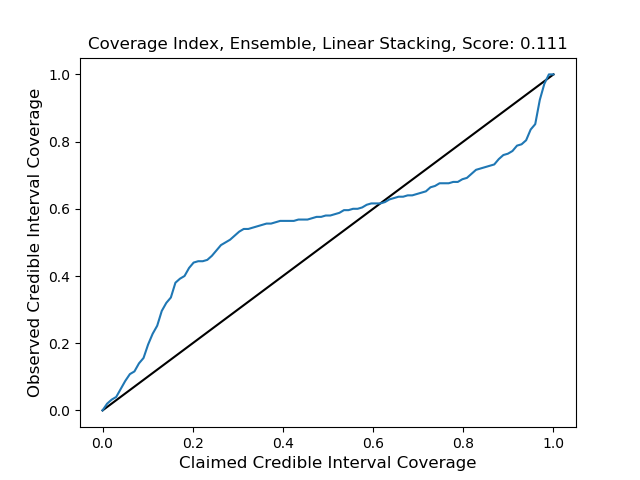}
  \includegraphics[width=.32\linewidth]{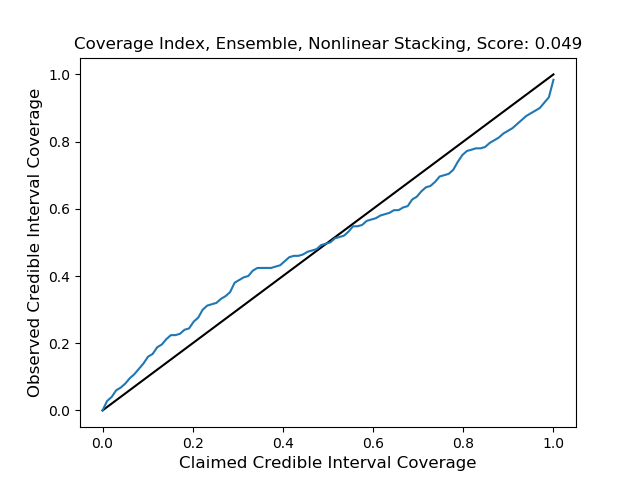}  
 \caption{Comparison in coverage probability of model's predictive interval of different ensemble methods. x-axis is the nomial coverage probability of model's predictive interval (in percentage), and y-axis is the actual coverage probability of model's predictive interval. Ideally, the coverage curve should align with the black line. For a given coverage percentage, curve below black line indicates underestimated uncertainty (i.e. overly narrow predictive interval), and curve above the black line indicates over-estimated uncertainty (i.e. unnecessarily wide predictive interval).
\\ 
 \textbf{Top Left}: Our model, with KL-only VI objective, \textbf{Top Right} Our model with KL+CvM VI objective. \textbf{Bottom}: (from left to right) \textbf{gam}, \textbf{lnr-stacking} and \textbf{nlr-stacking}.}
  \label{fig:exp_1d_comp_unc}
\end{figure}

We also quantitatively assessed \textbf{gam}, \textbf{lnr-stack}, \textbf{nlr-stack} and our model's quality in uncertainty quantification (in terms of the true coverage probability of the $p\%$ predictive intervals for $p \in (0, 1]$) in Figure \ref{fig:exp_1d_comp_unc}. In order to assess the effect of the calibrated VI objective \ref{eq:vi_obj}, we compare two versions of our model estimated with and without the CvM distance in VI objective. Compared to other ensemble methods, the nominal coverage of our model's predictive intervals are shown to be closer to their true coverage, and visible improvement can be observed for the model estimated with the CvM distance included in VI objective.

\clearpage
\section{Additional Figures for Real Data Application}

\begin{figure}[ht]
  \includegraphics[width=.33\linewidth]{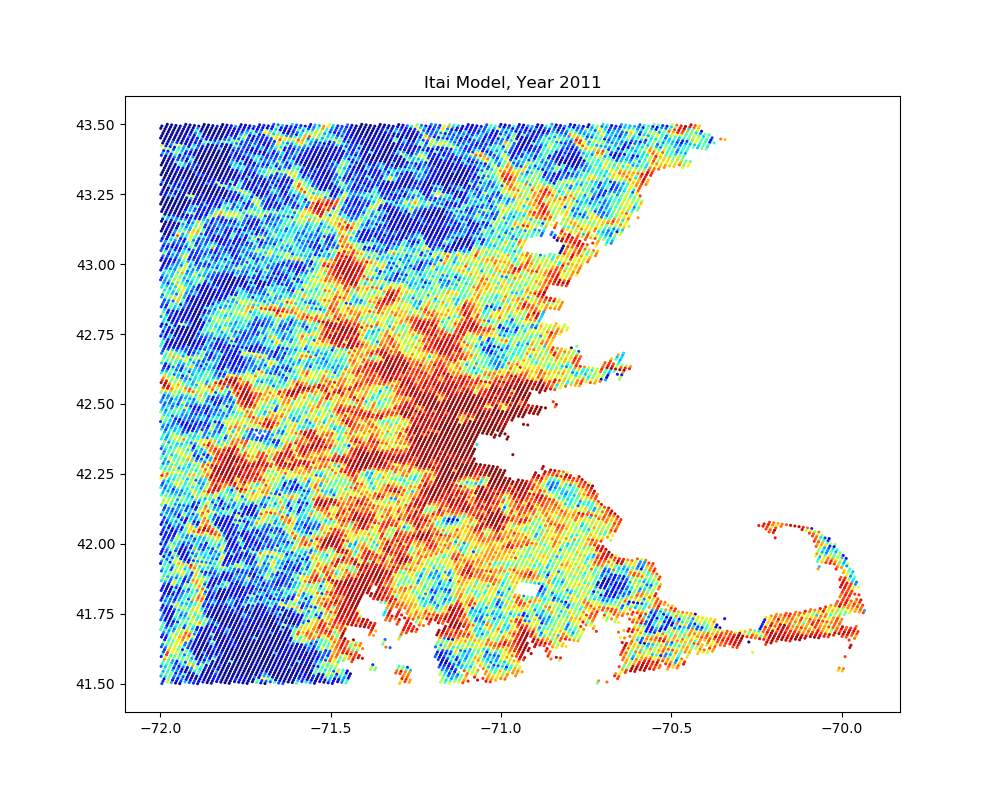}
  \includegraphics[width=.33\linewidth]{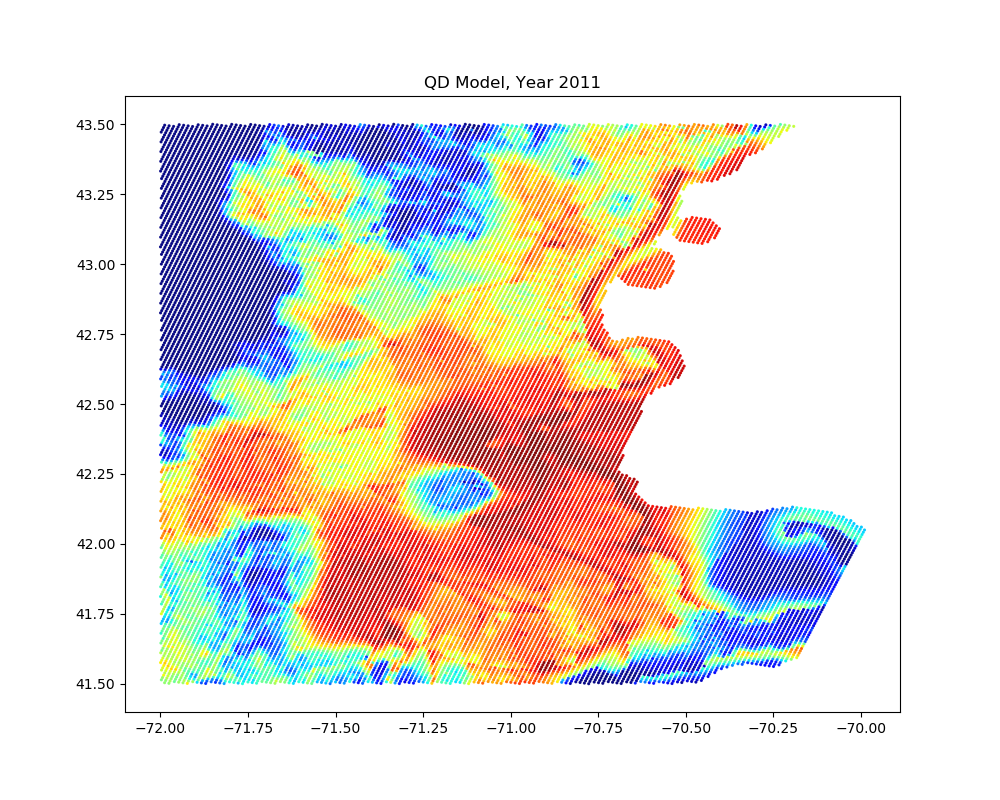}
  \includegraphics[width=.33\linewidth]{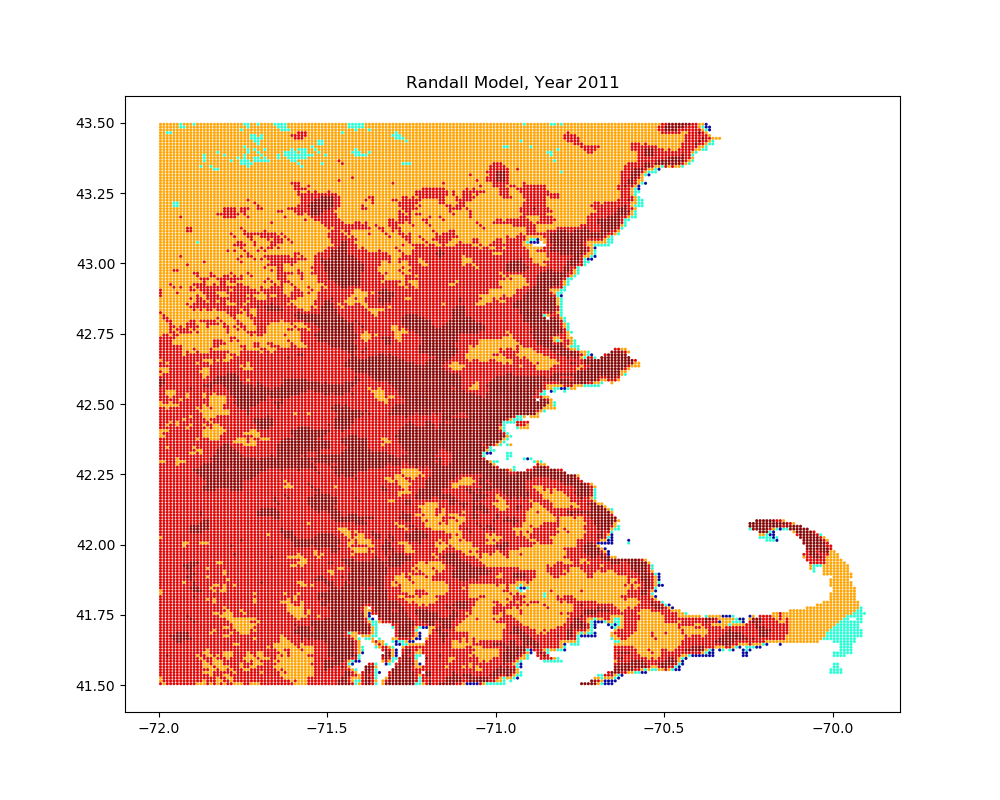}
\caption{Visualization of annual average $PM_{2.5}$ predictions from different base models in greater Boston region during Year 2011.}
\label{fig:pm25_base}
\end{figure}

\begin{figure}[!ht]
\centering
  \includegraphics[width=0.9\linewidth]{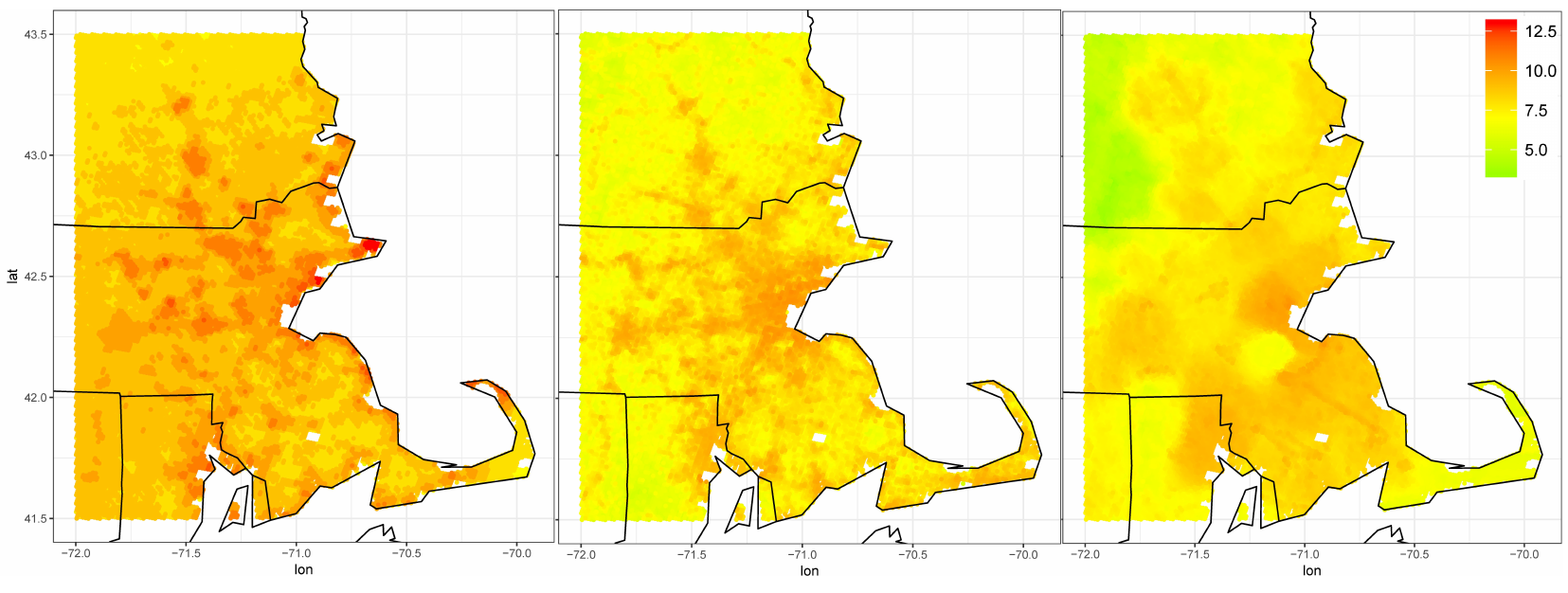}
  
  \includegraphics[width=.3\linewidth]{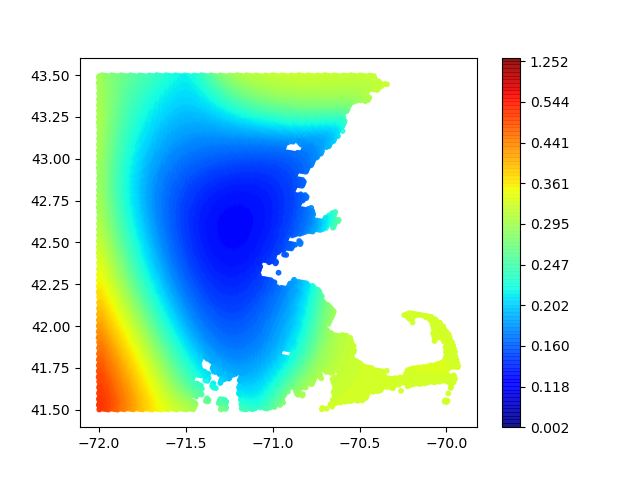}
    \includegraphics[width=.3\linewidth]{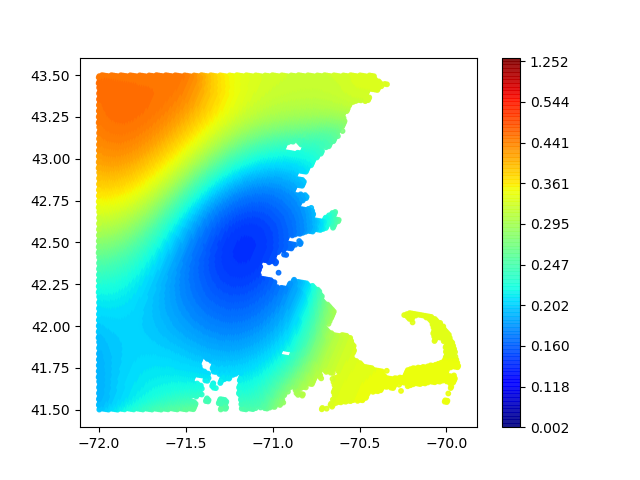}
  \includegraphics[width=.3\linewidth]{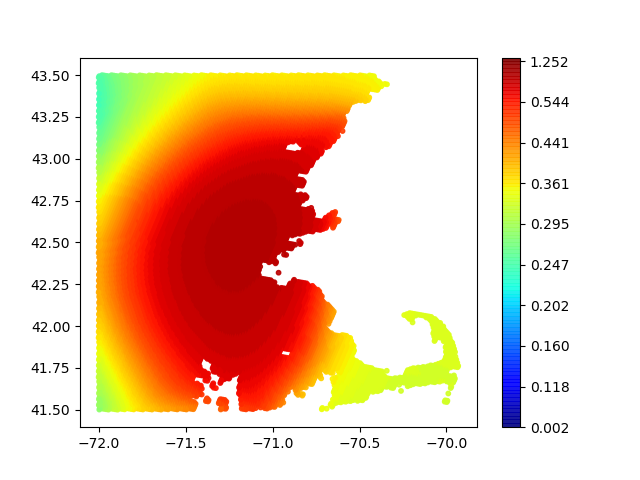}
  
\caption{\textbf{(Top)} Base model predictions and (Bottom) the posterior mean of the corresponding ensemble weight Gaussian Process.}
\label{fig:pm25_weight}
\end{figure}

\begin{figure}[!ht]
\centering
\scalebox{.8}{
\includegraphics[height=0.35\linewidth, width=0.7\linewidth]{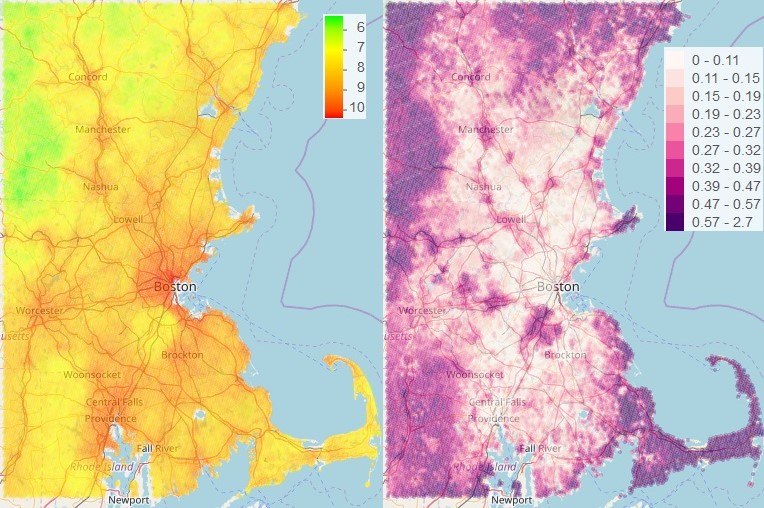}
}
\caption{Posterior predictive mean \textbf{(left)} and uncertainty (standard deviation) \textbf{(right)} for annual $PM_{2.5}$ in greater Boston area during year 2011. All unit in $\mu g/m^3$}
\label{fig:pm25_pred}
\end{figure}

\end{document}